\newif\ifarxiv
\arxivtrue
\documentclass[10pt]{article}

\ifarxiv
  \usepackage[preprint]{tmlr}
\else
  \usepackage{tmlr}
\fi

\usepackage[utf8]{inputenc}
\usepackage[T1]{fontenc}
\usepackage{hyperref}
\usepackage{url}
\usepackage{booktabs}
\usepackage{amsfonts}
\usepackage{amsmath}
\usepackage{nicefrac}
\usepackage{microtype}
\usepackage{xcolor}
\usepackage{graphicx}
\usepackage{multirow}
\usepackage{float}
\usepackage{orcidlink}

\title{How much does pretraining contamination help time series
foundation models?}

\author{\name Runshi Yang\,\orcidlink{0009-0005-7403-8399}\thanks{Corresponding author.} \email ryang@cs.toronto.edu \\
      \addr Horizon3 AI Labs, Unilever \\
      University of Toronto
      \AND
      \name Jyotsna Sehgal\,\orcidlink{0009-0009-0266-7232} \\
      \addr Work done while at Horizon3 AI Labs, Unilever
      \AND
      \name Khashayar Bayati\,\orcidlink{0009-0006-1837-0300} \email khashayar.bayati@torontomu.ca \\
      \addr Work done while at Horizon3 AI Labs, Unilever}

\ifarxiv
  \newcommand{\codewhere}{at \url{https://github.com/Runshi-Yang/tsfm-contamination}}
\else
  \newcommand{\codewhere}{in the supplementary material}
\fi
\newcommand{\codeavailability}{The code, the saved results and a script that rebuilds every table and
figure from them are \codewhere, together with \texttt{benchmark\_windows.csv} and
\texttt{exposure\_estimate.py}.}

\begin{document}

\maketitle

\newcommand{\nDirect}{299}
\newcommand{\stDirect}{258}
\newcommand{\nIndirect}{300}
\newcommand{\stIndirect}{282}
\newcommand{\corrIndirect}{0.630}
\newcommand{\nClean}{1055}
\newcommand{\stClean}{1055}
\newcommand{\dryAll}{73}
\newcommand{\verifiedDirect}{299}
\newcommand{\candidateDirect}{300}
\newcommand{\distMedian}{39}
\newcommand{\distNinety}{69}
\newcommand{\distMax}{2{,}299}
\newcommand{\gateDI}{0.535}
\newcommand{\gateDC}{0.602}
\newcommand{\gateIC}{0.607}
\newcommand{\maseSameLo}{97.3}
\newcommand{\maseSameHi}{99.2}
\newcommand{\maseZero}{0.6398}
\newcommand{\maseSmall}{0.6398}
\newcommand{\wqlMedian}{1.0000}
\newcommand{\wqlNear}{91}
\newcommand{\corrDCe}{0.441}
\newcommand{\rsqDCe}{0.19}
\newcommand{\corrDScrps}{0.997}
\newcommand{\rsqDScrps}{0.99}
\newcommand{\corrDMase}{1.000}
\newcommand{\rsqDMase}{1.00}
\newcommand{\maseDirect}{0.6802}
\newcommand{\maseIndirect}{0.6465}
\newcommand{\maseHarder}{5.2}
\newcommand{\ceIndirectCleanGap}{10.2}
\newcommand{\refSmall}{134.65}
\newcommand{\rawDiffSmall}{5.03}
\newcommand{\rawLoSmall}{\ensuremath{-5.65}}
\newcommand{\rawHiSmall}{16.05}
\newcommand{\rawBoundSmall}{4.2}
\newcommand{\adjDiffSmall}{3.15}
\newcommand{\adjLoSmall}{\ensuremath{-6.46}}
\newcommand{\rawWidthSmall}{21.7}
\newcommand{\adjWidthSmall}{19.4}
\newcommand{\adjBoundLo}{4.8}
\newcommand{\adjBoundHi}{5.2}
\newcommand{\rawBoundLo}{4.2}
\newcommand{\rawBoundHi}{4.6}
\newcommand{\adjGaps}{1.9, 2.3, 2.2 and 2.0}
\newcommand{\removedLo}{37.5}
\newcommand{\removedHi}{42.8}
\newcommand{\boundNatsSmall}{6.46}
\newcommand{\adjBoundSmall}{4.8}
\newcommand{\natsPerToken}{0.099}
\newcommand{\probGain}{10}
\newcommand{\scrpsAdjLo}{0.4}
\newcommand{\scrpsAdjHi}{1.2}
\newcommand{\scrpsAdjBySize}{0.4, 0.4, 0.7 and 1.2}
\newcommand{\scrpsRawLo}{6.3}
\newcommand{\scrpsRawHi}{7.1}
\newcommand{\blockTwoLo}{5.2}
\newcommand{\blockTwoHi}{5.7}
\newcommand{\blockFiveLo}{4.8}
\newcommand{\blockFiveHi}{5.1}
\newcommand{\nStationUnits}{540}
\newcommand{\nTwoDegUnits}{181}
\newcommand{\nFiveDegUnits}{48}
\newcommand{\expDay}{1.2}
\newcommand{\expBlend}{6}
\newcommand{\expExact}{0.0018}
\newcommand{\expOneIn}{551}
\newcommand{\drawsPerAug}{4.6}
\newcommand{\cpZeroTen}{0.31}
\newcommand{\cpOneTen}{0.45}
\newcommand{\pNinetyFiveSd}{0.011}
\newcommand{\robHorizstd}{2.1--2.5}
\newcommand{\robCtxmae}{0.8--1.4}
\newcommand{\robDryfrac}{3.0--3.3}
\newcommand{\rsqDry}{94}
\newcommand{\mdeAuc}{0.570}
\newcommand{\mdeGap}{11.5}
\newcommand{\didSd}{\ensuremath{-0.022}}
\newcommand{\didSdLo}{\ensuremath{-0.113}}
\newcommand{\didSdHi}{0.072}
\newcommand{\didPct}{\ensuremath{-1.1}}
\newcommand{\didPctLo}{\ensuremath{-5.4}}
\newcommand{\didPctHi}{3.4}
\newcommand{\didBound}{5.4}
\newcommand{\rawDidSd}{\ensuremath{-0.026}}
\newcommand{\rawDidSdLo}{\ensuremath{-0.122}}
\newcommand{\rawDidSdHi}{0.070}
\newcommand{\rawDidPct}{\ensuremath{-1.2}}
\newcommand{\rawDidPctLo}{\ensuremath{-5.8}}
\newcommand{\rawDidPctHi}{3.3}
\newcommand{\rawDidBound}{5.8}
\newcommand{\rsqTimesfm}{0.71}
\newcommand{\rsqOthers}{0.06--0.21}
\newcommand{\ctrlMean}{7.21}
\newcommand{\ctrlSd}{10.67}
\newcommand{\bonf}{0.0028}
\newcommand{\pExactTwenty}{0.036}
\newcommand{\outlierA}{54}
\newcommand{\outlierB}{88}
\newcommand{\seCellTwenty}{8.9}
\newcommand{\pExactForty}{\ensuremath{4.9{\times}10^{-4}}}
\newcommand{\exactMedLo}{\ensuremath{-1.0}}
\newcommand{\exactMedHi}{150.3}
\newcommand{\pBlendTwenty}{0.049}
\newcommand{\blendMeanTop}{18.1}
\newcommand{\pBlendTop}{0.07}
\newcommand{\relExcessTen}{2.2}
\newcommand{\relExcessTwenty}{17.0}
\newcommand{\moiraiPtHitDose}{320}
\newcommand{\moiraiPtHitCores}{four}
\newcommand{\moiraiPtHitP}{0.21}
\newcommand{\detLo}{0.48}
\newcommand{\detHi}{0.50}
\newcommand{\refBlendTop}{0.78}
\newcommand{\refExactTwenty}{0.89}
\newcommand{\probeTop}{2.68}
\newcommand{\probeHeld}{7.05}
\newcommand{\probeRatio}{2.6}
\newcommand{\probeDirect}{7.38}
\newcommand{\probeSdDirect}{0.25}
\newcommand{\probeIndirect}{7.11}
\newcommand{\probeSdIndirect}{0.06}
\newcommand{\probeClean}{7.04}
\newcommand{\probeSdClean}{0.19}
\newcommand{\probeSdHeld}{0.12}
\newcommand{\probeKone}{6.51}
\newcommand{\probeSdKone}{0.05}

\begin{abstract}
Time series foundation models are judged by their zero-shot accuracy on benchmark data, but their
pretraining corpora come from the same public archives as the benchmarks. A benchmark series may have
been seen in training (direct leakage), or series closely related to it may have been (indirect
leakage). We ask how large a score difference either is associated with. Comparing with data that is
certainly unseen does not answer this, since such data can only postdate the model and differs for
other reasons, so we take two other routes. First, on daily rain-gauge records whose presence in the
released Chronos training files we verify value by value, we compare in-corpus gauges with
out-of-corpus gauges from the same network and years. Across four Chronos checkpoints the in-corpus
windows are harder, not easier. After adjusting for a difference in intrinsic difficulty, the
membership-associated gap is below \adjBoundHi\% of token cross-entropy when stations are resampled and
below \blockTwoHi\% when nearby gauges are resampled together (one-sided 97.5\% limits, conditional on
the adjustment's assumptions). Five checkpoints from three families whose published corpora do not list
these gauges show the same gap, and as a control that uses no covariate they give a bound of
\rawDidBound\%. Four loss-based membership scores and the weight-movement signal of an adaptation method
do not detect membership on this benchmark. Second, we plant windows in a training corpus at known
doses and continue training six checkpoints from four families. Planted exact copies become detectable
after 8 to 80 exposures, depending on the checkpoint and its training schedule, and not at all in
Moirai; by our estimate from the published sampler, Chronos pretraining shows each day of these records
in about one training crop and almost never a whole window. TimesFM is the only family whose planted
near-duplicates and blends both pass the share test, and Chronos shows a weak blend signal at 320
exposures. On the benchmark we could audit, the membership-associated gap is small.
\end{abstract}

\section{Introduction}
\label{sec:intro}

A time series foundation model (TSFM) is pretrained once, on a large collection of series from
many domains, and then asked to forecast series it has never been trained on. Chronos
\citep{ansari2024chronos}, TimesFM \citep{das2024timesfm}, Moirai \citep{woo2024moirai} and
Lag-Llama \citep{rasul2023lagllama} all work this way. Because such a model is not fitted to the
series it forecasts, the only way to judge it is to run it on benchmark data, and benchmark scores
are what people use to decide which model to deploy.

That judgement is fair only if the benchmark data is new to the model. Pretraining corpora are
assembled from public archives, and benchmarks are drawn from the same archives, so the assumption
can fail in two ways \citep{meyer2025rethinking}. In \emph{direct leakage}, a benchmark series was
itself part of the pretraining corpus. In \emph{indirect leakage}, the benchmark series was not, but
closely related series were: another sensor that recorded the same weather on the same days, or a
reanalysis product built from the same observations. Either could make a model look better than it
would on genuinely new data. \citet{meyer2025rethinking} describe both and leave their size open.

The usual response is detection: decide, for a given example, whether it was in the training set
\citep{shokri2017membership}. For language models, loss-based detectors have been proposed for
pretraining data \citep{shi2024detecting}, but \citet{duan2024membership} find that membership
inference against large language models barely beats random guessing in most settings, partly
because each example is seen about once, and \citet{das2025blind} show that baselines which never
look at the model can beat published attacks, because members and non-members often differ in ways
unrelated to training. For TSFMs, \citet{li2026tsfmaudit} audit whether whole datasets were in a
corpus by watching how a model adapts when briefly fine-tuned on them. We ask a different question:
by how much could leakage have changed the score? Following \citet{magar2022contamination}, who
separate whether contaminated data was memorised from whether it improved the downstream score, we
measure the second quantity, and answer with an upper confidence limit rather than a verdict. A
detector that finds nothing leaves open that the effect is large and simply invisible to it; a
bound says how large the effect could be given what was observed.

Effect sizes have been measured before. In language models, \citet{schaeffer2026quantifying} pretrain
models with controlled numbers of test-set replicas and find that even a single replica lowers test
cross-entropy, so a small effect here is a finding about this setting, not a general rule. For TSFMs,
\citet{aksu2024gifteval} retrain Moirai on a corpus without their benchmark's training splits and find
that the original, leaking Moirai scores better on most affected datasets, increasingly so at longer
horizons; \citet{meyer2025rethinking} reproduce that comparison and, for indirect leakage, train a
small transformer with and without temporally overlapping correlated series and find a mean change in
error of about 12\%. Both compare whole training corpora, report no interval, and in the second case do
not use a released foundation model. Our measurement differs in three ways. Membership is verified window
by window against a released model's training files; the quantity is an upper confidence limit on a
released checkpoint, without retraining it, and conditional on stated assumptions; and the planted
experiments give exposure thresholds, for direct and indirect forms, across four families.

The obvious experiment compares a model's accuracy on leaked data with its accuracy on data that was
certainly not in the corpus. The only data certain to be absent is data recorded after the model was
released, and newer data differs from older data for reasons unrelated to leakage. On our benchmark
the post-release group is easier to forecast by about twice the effect we want to measure, and a
check that uses no model at all can already tell it apart from the older data
(Section~\ref{sec:gate}). We therefore take two routes that avoid that comparison. The first is
observational. NOAA's Global Historical Climatology Network archives daily precipitation from U.S.\
gauges; one subset of them is in the Chronos pretraining corpus and neighbouring gauges from the
same network are not. We verify membership value by value and compare the two groups over the same
years. The second is interventional. We plant windows in a training corpus a known number of times,
as exact copies, near-duplicates or blends with other series, continue training, and find how many
exposures it takes before the model fits a planted window measurably better than one it never saw.
The first route gives the bound; the second explains why the bound is small, and shows when it would
not be. In plain terms, the first study checks whether a released model forecasts the rain gauges it
was trained on better than similar gauges it never saw. The second puts copies of series into a
model's training data ourselves and counts how many times the model has to see a series before it
forecasts that series noticeably better.

Our contributions are:
\begin{itemize}
\item \textbf{A bound on the membership-associated gap.} On four released Chronos checkpoints from
20M to 709M parameters, in-corpus windows are harder to forecast, not easier. After adjusting for a
model-free measure of difficulty, the membership-associated gap is below \adjBoundLo--\adjBoundHi\%
of token cross-entropy (one-sided 97.5\% limit), and below \blockTwoLo--\blockTwoHi\% when nearby
gauges are resampled together. Five checkpoints from three other families serve as a covariate-free
control and bound the gap at \rawDidBound\% (Section~\ref{sec:bound}).
\item \textbf{A benchmark with verified membership.} \verifiedDirect\ of \candidateDirect\ in-corpus
windows are matched value by value against the released corpus; every comparison is screened with a
check that uses no model; and the two accuracy metrics usually reported for these models, MASE and
WQL, turn out to be nearly insensitive to the model on this data (Section~\ref{sec:benchmark}).
\item \textbf{Exposure thresholds, across four families.} Planted exact copies become detectable in
the 46M Chronos model after 20 exposures (40 by the mean test alone). Across six checkpoints from four
families, each under its own continued-training schedule, the threshold ranges from 8 exposures to 80,
and Moirai shows none up to 320. We
estimate from the published Chronos sampler that pretraining shows each day of a USHCN record in about
\expDay\ unmixed training crops and a whole window almost never. TimesFM is the only family whose
planted near-duplicates and blends both pass the share test (Section~\ref{sec:detect}).
\item \textbf{Four loss-based scores and one movement probe, silent on the benchmark and responsive
on planted data.} The four scores give AUROCs between \detLo\ and \detHi\ on the benchmark, and the
weight-movement signal of an adaptation method does not separate the groups; all of them separate
planted exact copies from unseen windows (Section~\ref{sec:detectors}).
\end{itemize}

The observational bound covers one model family and one variable. It bounds the gap associated with
having seen these particular gauges on these particular days, not the benefit of general familiarity
with American weather, which both groups share (Section~\ref{sec:scope}). We do not claim that
leakage is harmless for TSFMs in general, and the cross-family results give a reason not to.

\section{The benchmark and protocol}
\label{sec:benchmark}

\subsection{Three groups of windows}
\label{sec:groups}

We use daily precipitation from GHCN \citep{menne2012ghcn}. Stations belonging to the U.S.\
Historical Climatology Network (USHCN) \citep{menne2009ushcn} appear in the published pretraining
corpus of Chronos \citep{ansari2024chronos}. The other stations we use are \emph{co-operative}
stations: gauges in the same national network, reporting the same variable through the same archive,
which do not appear in that corpus.

A \emph{window} is 576 consecutive days from one station, split into 512 days of context and a
64-day horizon to be forecast; these match the context and prediction lengths of the
\texttt{chronos-t5} models. A window is kept only if all 576 days are present and none was flagged by
NOAA's quality control. We build three groups (Table~\ref{tab:pool}).

\begin{itemize}
\item \textbf{direct}: windows from USHCN stations, 2010--2014. These are in the corpus.
\item \textbf{indirect}: windows from co-operative stations over the same years. These gauges are
not in the corpus, but the weather they recorded reaches it through other routes (below).
\item \textbf{clean}: windows from the same co-operative stations, starting June 2024, after the
checkpoints were built. These cannot be in the corpus.
\end{itemize}

\begin{table}[t]
\centering
\small
\caption{The evaluation pool. \texttt{corpus\_corr} is the median over windows of the largest
correlation between a window and the records of the five nearest corpus stations over the same
dates; it is $1$ for \textbf{direct} by definition and undefined for \textbf{clean}, since the corpus
ends before 2024. Dry days is the share of days with no precipitation.}
\label{tab:pool}
\begin{tabular}{llrrrr}
\toprule
group & period & windows & stations & \texttt{corpus\_corr} & dry days \\
\midrule
direct & 2010--2014 & 299 & 258 & 1.000 & 0.708 \\
indirect & 2010--2014 & 300 & 282 & 0.630 & 0.721 \\
clean & 2024-06 onward & 1055 & 1055 & --- & 0.736 \\
\bottomrule
\end{tabular}

\end{table}

The comparison that carries the bound is \textbf{direct} against \textbf{indirect}. The two groups
come from the same network, the same years and the same processing code, and occupy the same
territory: the median \textbf{indirect} station lies \distMedian\,km from the nearest corpus station
(90th percentile \distNinety\,km), well within the size of a weather system. What differs is whether
the gauge's own record was in training.

\textbf{indirect} is not independent of the corpus, which is why it is not called clean. ERA5
reanalysis \citep{hersbach2020era5}, which is in the Chronos corpus, describes the atmosphere on the
same days on a coarse grid, and nearby USHCN gauges recorded the same storms; the median correlation
of \corrIndirect\ in Table~\ref{tab:pool} measures this overlap. So \textbf{indirect} carries
indirect leakage by construction, and the \textbf{direct}--\textbf{indirect} gap isolates what the
gauge's own record added on top of it.

\subsection{Verifying membership}
\label{sec:verify}

A dataset appearing in a corpus listing does not guarantee that a particular window was trained on.
We matched each \textbf{direct} window day by day against the \texttt{ushcn\_daily} rows of the
released Chronos training data. The two archives store precipitation in different units (tenths of a
millimetre and hundredths of an inch); after converting, \verifiedDirect\ of the \candidateDirect\
windows match exactly. The remaining window has one day missing from the released file and is
excluded everywhere. This proves that the windows are present in the released training files, not that
the training sampler drew them; how often it would have is estimated in Section~\ref{sec:family}. We
also checked that the released data contains no second slice of U.S.\ co-operative station records,
so the \textbf{indirect} gauges' own measurements are absent. The two checks differ in strength:
presence of \textbf{direct} is verified at the level of values, absence of \textbf{indirect} at the
level of the dataset listing.

\subsection{A check that uses no model}
\label{sec:gate}

Whether a dataset ends up in a corpus is often caused by properties of the dataset, so two groups
can differ for reasons unrelated to training \citep{das2025blind}. Before using any comparison we
ask whether the groups can be told apart without a model. We compute simple descriptors of each raw
window (spectral band powers, first-difference statistics, lag-one autocorrelation, spectral entropy
and the dry-day fraction) and require that no single descriptor separates the groups with an AUROC
above $0.60$. The AUROC (area under the receiver operating characteristic curve) is the probability
that a randomly chosen window from one group scores higher than one from the other, so $0.5$ is
chance. The threshold was fixed before any model was run.

\textbf{direct} against \textbf{indirect} passes, at \gateDI. Both comparisons involving
\textbf{clean} fail, at \gateDC\ (\textbf{direct}) and \gateIC\ (\textbf{indirect}), on the same
spectral band: the same stations recorded differently shaped rainfall a decade later, and a
model-free descriptor can see it. The model sees it too. For the 46M checkpoint, \textbf{clean}
windows have a cross-entropy \ceIndirectCleanGap\ nats lower than \textbf{indirect} windows from the
same pool of stations, about twice the \textbf{direct}--\textbf{indirect} difference. A comparison
against post-release data would mostly measure the change between decades, which is why we do not
make it.

Passing the check does not settle the matter. It screens one descriptor at a time, and with about
270 stations per group the smallest AUROC this design can reliably distinguish from chance is
\mdeAuc\ (Section~\ref{sec:power}), so a pass at \gateDI\ means only that no large single-feature
difference exists. Section~\ref{sec:confound} describes a difference between \textbf{direct} and
\textbf{indirect} that the check missed and that we correct for.

\subsection{Metrics, and why two of them are nearly insensitive here}
\label{sec:metric}

Chronos converts a series into discrete tokens and is trained as a language model over them, so its
natural loss is token cross-entropy. For each window we feed the true horizon to the model (teacher
forcing) and sum the cross-entropy over the 64 horizon tokens and the end-of-sequence token, in
nats. We call this CE; lower is better, and percentages of CE are percentages of this sum. We score
four released checkpoints: \texttt{mini} (20M parameters), \texttt{small} (46M), \texttt{base} (201M)
and \texttt{large} (709M).

We also computed the two metrics the Chronos evaluation reports: the mean absolute scaled error
(MASE), the mean absolute error of the median forecast divided by the mean day-to-day change of the
context, and the weighted quantile loss (WQL), the pinball loss averaged over nine quantiles and
divided by the total observed rain. On this data both are nearly insensitive to the model. About
\dryAll\% of days are dry, so
the median of the model's sampled forecasts is zero on most horizon days. MASE then reduces to a
property of the observed rain alone: an all-zero forecast implies a mean MASE of \maseZero\ over the
pool, and the 46M model's measured value is \maseSmall. Comparing the 20M model with each larger one,
MASE is exactly identical on \maseSameLo\ to \maseSameHi\% of windows. WQL moves only in the third
decimal place across model sizes; its median over windows is \wqlMedian, its value for an all-zero
forecast. Scaled CRPS does respond to the model, but it correlates with the model-free difficulty
measure of Section~\ref{sec:confound} at \corrDScrps, so its adjusted bound of \scrpsAdjLo--\scrpsAdjHi\%
says little about the model (Appendix~\ref{app:metrics}). We make CE the primary metric because
it is the model's own training objective and, of the four, the one least dominated by properties of
the windows (it correlates with the difficulty measure at \corrDCe), and state all results in it. The lesson is general and cheap
to apply: before quoting a metric, check that it changes when the model does.

\subsection{Models}
\label{sec:models}

We use nine released checkpoints from four families (Table~\ref{tab:models}). Only Chronos has the
USHCN gauges in its corpus. For the other three families we checked the published corpus
descriptions, at the level of dataset listings: LOTSA, Moirai's corpus, has no USHCN or GHCN subset
among its 170; the Lag-Llama checkpoint records its twenty training sets and none is a rain-gauge
archive; and the public sources TimesFM lists do not include one. As far as their published corpora
show, those five checkpoints see \textbf{direct} and \textbf{indirect} as equally unfamiliar, and they
serve as controls. The two TimesFM checkpoints are two versions trained differently (1.0 with 200M
parameters, 2.0 with 500M), not two sizes of one model. Every model is
scored with its own training loss on the 64-day horizon, since that is where memorisation would show:
token cross-entropy for Chronos, the negative log-likelihood of the predictive distribution for
Moirai and Lag-Llama, and for TimesFM, which has no likelihood, the squared error of its mean forecast
plus the pinball loss of its nine quantiles, after dividing by the context's standard deviation.
Section~\ref{sec:detect} uses six of the checkpoints for planted-data experiments; the last column of
Table~\ref{tab:models} anticipates the learning rates chosen there.

\begin{table}[t]
\centering
\small
\caption{The nine checkpoints. On the rain-gauge benchmark, Chronos is the only family whose corpus
contains the \textbf{direct} gauges; the others are controls. The last column is the learning rate
used for the planted-data experiments of Section~\ref{sec:detect}, chosen by a rule that sees only
background windows; Moirai large could not be trained stably at any rate in the grid.}
\label{tab:models}
\begin{tabular}{lllll}
\toprule
family & checkpoint & training loss & USHCN in corpus & planted-data learning rate \\
\midrule
Chronos & 20M & token cross-entropy & yes & not planted \\
 & 46M &  & yes & $10^{-4}$ \\
 & 201M &  & yes & not planted \\
 & 709M &  & yes & $10^{-4}$ \\
\midrule
Moirai 1.0-R & small & NLL (mixture) & no & $10^{-5}$ \\
 & large &  & no & no stable rate \\
\midrule
TimesFM & 1.0-200M & squared + pinball & no & $10^{-5}$ \\
 & 2.0-500M &  & no & $10^{-5}$ \\
\midrule
Lag-Llama & one size & NLL (Student-$t$) & no & $10^{-4}$ \\
\bottomrule
\end{tabular}

\end{table}

\section{The bound}
\label{sec:bound}

\subsection{The raw gap}
\label{sec:raw}

For each checkpoint we take the difference in mean CE,
\[
  \Delta = \overline{\mathrm{CE}}_{\text{direct}} - \overline{\mathrm{CE}}_{\text{indirect}}.
\]
If having seen a window made it easier to forecast, $\Delta$ would be negative. It is positive at
every size (Table~\ref{tab:bound}): the in-corpus windows are the harder ones.

To say how large a hidden advantage could be, we put a 95\% interval on $\Delta$ by resampling
stations with replacement 10{,}000 times; two windows from one gauge share a climate and an
instrument, so the station is the independent unit \citep{cameron2008bootstrap}. We then take the end
of the interval most favourable to leakage and divide it by the \textbf{indirect} mean. We call the
result the \emph{bound}: the largest gap in favour of the in-corpus windows that is still consistent
with the data. Because it is one end of a two-sided 95\% interval, it is a one-sided 97.5\% upper
confidence limit. For the 46M model the interval is $[\rawLoSmall, \rawHiSmall]$ nats against an
\textbf{indirect} mean of \refSmall, so the bound is \rawBoundSmall\%. The raw bounds for all four
sizes lie between \rawBoundLo\ and \rawBoundHi\%.\footnote{All gaps and bounds are computed from
unrounded per-window values, so they can differ in the last digit from differences of the rounded
group means in Appendix~\ref{app:metrics}.}

Resampling stations treats different gauges as independent, but nearby gauges record the same storms:
the median \textbf{indirect} station is \distMedian\,km from a corpus station. Resampling whole
$2^\circ$ by $2^\circ$ cells instead, so that gauges sharing weather are drawn together, widens the
intervals and raises the adjusted bound to \blockTwoLo--\blockTwoHi\% (Appendix~\ref{app:blockboot}).
We report the station-level figure as the main result, with the $2^\circ$ figure beside it as the
more conservative one.

The comparison is observational. The \textbf{direct} gauges were not assigned to the corpus at
random, so what we bound is the gap \emph{associated} with membership, not a causal effect of it.
The next two subsections remove the part of that gap we can attribute to other differences between
the groups, one with a model-free covariate and one with models that never saw the gauges.

\subsection{The two groups differ in difficulty}
\label{sec:confound}

The raw comparison assumes that \textbf{direct} and \textbf{indirect} windows are equally hard to
forecast. They are not. MASE, which Section~\ref{sec:metric} showed to be nearly insensitive to the
model on this data, is \maseHarder\% higher for \textbf{direct} (\maseDirect\ against
\maseIndirect) before any model is involved. We turn this into a covariate. For each window,
\[
  d = \frac{\mathrm{mean}_t |y_t|}{\mathrm{mean}_t |Y_t - Y_{t-1}|},
\]
with the numerator over the 64-day horizon and the denominator over the 512-day context. This is the
MASE an all-zero forecast would receive: how much rain there is to predict, relative to how much the
series usually moves. It uses only the data, and it explains $R^2 = \rsqDCe$ of the variation in CE
across windows.

Adjusting for $d$ corrects the comparison if $d$ is a common cause of group membership and of CE, and
not a consequence of membership. The second is impossible: whether a gauge's record entered a
training corpus assembled a decade later cannot change the rain it measured in 2012. The first is
plausible. USHCN was selected for long, high-quality records and even spatial coverage, and that
selection may well favour gauges whose rainfall is harder to forecast. We therefore estimate the
coefficient $\beta$ in
\[
  \mathrm{CE} = \alpha + \beta\,\mathbf{1}[\text{direct}] + \gamma\, d + \varepsilon
\]
by least squares, with the same bootstrap for its interval. The adjustment rests on three
assumptions, and every adjusted bound below is conditional on them: that CE depends on $d$ linearly, as
the model says; that the two groups overlap in $d$, so the comparison is not an extrapolation; and that
no confounder other than $d$ separates them.

\begin{table}[t]
\centering
\small
\caption{\textbf{direct} minus \textbf{indirect} token cross-entropy, before and after adjusting for
difficulty. $\Delta$ and $\beta$ are in nats summed over the horizon; intervals are 95\%, resampling
stations. The bound is the largest membership-associated gap consistent with the data, as a
percentage of the \textbf{indirect} mean (one-sided 97.5\%).}
\label{tab:bound}
\begin{tabular}{lrrcrrr}
\toprule
& \multicolumn{2}{c}{raw} & & \multicolumn{3}{c}{adjusted for difficulty} \\
\cmidrule(lr){2-3}\cmidrule(lr){5-7}
checkpoint & $\Delta$ & bound & & $\beta$ & 95\% interval & bound \\
\midrule
20M & $+$4.52 & 4.6\% & & $+$2.59 & $[-7.11,\ 12.46]$ & \textbf{5.2\%} \\
46M & $+$5.03 & 4.2\% & & $+$3.15 & $[-6.46,\ 12.89]$ & \textbf{4.8\%} \\
201M & $+$4.66 & 4.4\% & & $+$2.82 & $[-6.32,\ 12.11]$ & \textbf{4.9\%} \\
709M & $+$4.39 & 4.4\% & & $+$2.57 & $[-6.33,\ 11.69]$ & \textbf{5.0\%} \\
\bottomrule
\end{tabular}

\end{table}

\subsection{The bound}
\label{sec:headline}

Adjustment removes between \removedLo\ and \removedHi\% of the raw gap, depending on the checkpoint.
The estimate stays positive at every size, every interval still contains zero, and the bound moves
from \rawBoundLo--\rawBoundHi\% to \adjBoundLo--\adjBoundHi\% (Table~\ref{tab:bound}). The bound grows while the estimate shrinks because adjustment moves the
whole interval towards zero: at 46M the interval narrows from \rawWidthSmall\ to \adjWidthSmall\
nats wide, but its centre drops from \rawDiffSmall\ to \adjDiffSmall\ and its lower end moves from
\rawLoSmall\ to \adjLoSmall. Our main result is therefore:

\begin{quote}
On GHCN daily precipitation, for released \texttt{chronos-t5} checkpoints from 20M to 709M
parameters, the difficulty-adjusted membership-associated gap in token cross-entropy is below
\adjBoundLo\ to \adjBoundHi\% of the out-of-corpus mean when stations are resampled, and below
\blockTwoLo\ to \blockTwoHi\% when nearby gauges are resampled together, as one-sided 97.5\% confidence
limits conditional on the assumptions of the adjustment.
\end{quote}

A percentage of a summed cross-entropy is hard to picture, so it helps to translate one checkpoint's
bound. For 46M the bound is \adjBoundSmall\%, or \boundNatsSmall\ nats over the 65 scored tokens (64
forecast days and the end-of-sequence token), which is \natsPerToken\ nats per scored token. A gap that
large would mean the model puts about \probGain\% more probability on the observed values, as a
geometric mean over the horizon rather than on every day. This is the most the data allow; the measured
gap points the other way.

The bound shows no trend with model size, and neither does the adjusted gap, which is \adjGaps\% of
the reference mean from 20M to 709M. Over a 35-fold range of parameters we see no sign that larger
models draw more from membership. The four checkpoints share a corpus and are scored on the same
windows, so they are not four independent replications.

\paragraph{Does the result depend on how difficulty is defined?} We repeated the adjustment with
three other model-free definitions: the share of dry days in the horizon, the standard deviation of
the horizon, and the day-to-day variability of the context alone (Appendix~\ref{app:robust}). They
give bounds of \robDryfrac\%, \robHorizstd\% and \robCtxmae\%. The dry-day share explains \rsqDry\%
of the variation in CE by itself, so the linearity and overlap assumptions carry more weight for it
than for the others, but it is not excluded. Each figure is conditional on the same three assumptions
for its own covariate. The definition we report gives the loosest bound of all four, and the estimated
gap is positive under every one.

\subsection{The same comparison on every checkpoint}
\label{sec:did}

The covariate removes the part of the gap that $d$ can see. The five checkpoints whose published
corpora do not list the USHCN gauges (Section~\ref{sec:models}) offer a check that needs no covariate:
for them
\textbf{direct} and \textbf{indirect} are equally out of corpus, so their gap contains everything that
separates the two sets of gauges except membership. If membership were associated with an advantage
in Chronos, the Chronos gaps should be more negative than theirs. Each model is scored with its own
loss, and negative log-likelihoods have no natural zero, so percentages of the mean do not compare
across models. We therefore standardise each model's per-window loss by its standard deviation over
the two groups (Table~\ref{tab:did} and Figure~\ref{fig:step1} in Appendix~\ref{app:did}).

Every checkpoint finds the USHCN windows harder, including the five that never saw them, and the four
Chronos gaps sit within the range of the others. The harder \textbf{direct} windows are therefore a property
of the gauges, not of training. Without any covariate, the difference between Chronos and the average
of the other models is \rawDidSd\ standard deviations, with an interval from \rawDidSdLo\ to
\rawDidSdHi; in Chronos's own units this is \rawDidPct\% of the 46M model's \textbf{indirect} mean CE,
with an interval from \rawDidPctLo\% to \rawDidPctHi\%. This is the independent check: it shares
no assumption with the adjusted bound beyond the resampling, and the largest gap it allows,
\rawDidBound\%, is close to the \adjBoundLo--\adjBoundHi\% obtained with the covariate. The point
estimate lies on the leakage side, but it is small and the interval contains zero. Adding the covariate
to this comparison changes little (\didSd\ standard deviations, bound \didBound\%).

This control rests on its own assumption: that the other architectures respond to what makes a gauge
hard in the same way Chronos does, once losses are standardised. TimesFM's loss tracks $d$ far more
closely than the others' ($R^2 = \rsqTimesfm$ against \rsqOthers), one sign that they do not all
respond alike. The controls' absence from their corpora is also established only at the level of
dataset listings. We therefore treat this as a second check on the bound, not a replacement for it.

\subsection{How sharp is the study?}
\label{sec:power}

A bound is not the smallest effect the study could detect. Using the same station-clustered standard
error, a true gap would have to be \mdeGap\% of the \textbf{indirect} mean, at every checkpoint size,
for this design to find it with 80\% power at two-sided $\alpha = 0.05$. For a membership detector
scored by AUROC, the corresponding figure is \mdeAuc\ \citep{hanley1982meaning}. The bound is
tighter than the resolution because the observed difference fell on the side opposite to leakage. The
accurate pair of statements is that the membership-associated gap is below about 5 to 6\%, and that
the
study would reliably have detected a true gap of \mdeGap\% or more. Gaps between the two are not ruled
out by power, only by the observed direction.

\subsection{What the bound covers, and what it cannot}
\label{sec:scope}

Both groups are U.S.\ gauges from the same years. We assume that whatever the model learned about
American weather in general benefits both groups equally and so cancels from the difference; the
shared country and period make this plausible but do not prove it. The \textbf{direct} windows are corpus
records; the \textbf{indirect} ones correlate with contemporaneous corpus records at a median of
\corrIndirect. The bound therefore covers the step from the second to the first: having seen this
gauge on these days, over and above having seen the same storms through other instruments. It does
not cover general familiarity with the domain, and no comparison of this kind can, because any
control would come from the same public archives the model was trained on. That part of the question
can only be handled when an evaluation set is built. Whether it should count as leakage at all is
debatable, since learning the domain is what pretraining is for; the part that is unambiguously a
problem, having seen the exact realisation being evaluated, is the part the bound covers.

\section{Why nothing is detectable here, and when it would be}
\label{sec:detect}

The bound says how large a membership-associated gap could be. It does not say why the gap is
small. This section asks the detection question directly, then creates leakage at known doses to
find out how much exposure a model needs before any trace of it appears, in all four families.

\subsection{Detectors on the benchmark}
\label{sec:detectors}

We ran four loss-based membership scores and one probe on \textbf{direct} against
\textbf{indirect}, for the two Chronos checkpoints we also use below. The four scores are the plain loss,
Min-20\%, which averages the 20\% of horizon tokens the model finds least likely
\citep{shi2024detecting}, the loss after regressing out the difficulty measure $d$, and the loss
relative to a reference model whose published corpus does not list the gauges (Moirai small,
standardised). The probe takes one signal from the adaptation method of \citet{li2026tsfmaudit}: we
fine-tune briefly on a group of windows and record how far the weights have to move,
$\lVert\Delta\theta\rVert / \lVert\theta\rVert$, on the reasoning that data a model already knows needs
less change. Their method combines several adaptation features over whole datasets, and their own
analysis ranks movement alone as its weakest feature; we use only movement, so this is not a test of
their method. We fine-tune for 120 steps on ten windows per group, over five seeds; each seed draws a
different ten windows from the group and a different optimisation order.

None of them detects membership (Table~\ref{tab:detect} in Appendix~\ref{app:detect}). The four loss-based scores give AUROCs
between \detLo\ and \detHi\ on both checkpoints, every interval contains $0.5$, and all are below the
\mdeAuc\ that the design could resolve (Section~\ref{sec:power}). The probe needs $\probeDirect \pm
\probeSdDirect$, $\probeIndirect \pm \probeSdIndirect$ and $\probeClean \pm \probeSdClean$ ($\times
10^{-3}$, mean $\pm$ sd over seeds) of relative weight movement for \textbf{direct}, \textbf{indirect}
and \textbf{clean}; the groups overlap within one seed standard deviation, and \textbf{direct}, the
in-corpus group, needs the most. The unit scored by the probe is a group of ten windows, not a window,
so it yields no membership AUROC; that would need independent labelled audit examples. A silent
detector means little unless it speaks when there is something to find, so we now plant leakage and
point the same detectors at it.

\subsection{Planting leakage at a known dose}
\label{sec:inject-design}

We build a training corpus from GHCN precipitation windows of the same shape as the benchmark, drawn
from 5{,}932 stations that share none with the evaluation pool, so this experiment cannot contaminate
Section~\ref{sec:bound}. It holds 8{,}000 background windows and 180 \emph{core} windows, each planted
a known number of times $k \in \{1, 2, 4, 8, 16, 32\}$ in one of three forms:
\begin{itemize}
\item \textbf{exact}: identical copies of the core, the strongest possible direct leakage;
\item \textbf{near-duplicate}: copies correlated with the core at $\rho = 0.99$, with the same spectrum
and the same distribution of values, so the core itself never appears;
\item \textbf{blend}: convex combinations of the core with a randomly chosen other window, the core's
weight between $0.3$ and $0.7$, which is the kind of series Chronos's TSMixup augmentation produces.
\end{itemize}
The last two are indirect leakage: the model sees series closely related to the core, never the core.
With ten cores per combination of form and $k$, the planted copies are $19.1\%$ of a corpus of
$9{,}890$ windows, and 400 further windows from the same stations are never planted and serve as the
control. All eighteen combinations sit in one corpus, and each checkpoint is trained on it once, for
ten epochs with batch size 16, so differences between cells cannot come from differences between
runs. Every training example is a whole 576-day window, so one exposure here means the exact window
presented once; a window planted $k$ times is seen $10k$ times, and we report doses in these exposures.
Pretraining exposure is a different unit, and Section~\ref{sec:family} relates the two.

\paragraph{What counts as memorised.} Being planted is not the same as being memorised. For every
window $z$ we measure how much better the model fits it after training,
$\delta(z) = \mathrm{loss}_{\text{before}}(z) - \mathrm{loss}_{\text{after}}(z)$. Training improves the
model on all rainfall, so the never-planted controls improve too (for Chronos 46M, by \ctrlMean\ nats
on average, standard deviation \ctrlSd); a planted window counts as memorised only to the extent that
it improves by more. We judge each cell in two ways, both Bonferroni-corrected over the eighteen cells
($\alpha = \bonf$). The \emph{share test} counts how many of the ten cores improve by more than the
controls' 95th percentile; each does so by chance with probability $0.05$, and four or more
($p = 0.001$) passes. The percentile is itself estimated from 400 controls, so the chance rate is
$0.05$ only up to a sampling error of about \pNinetyFiveSd. The \emph{mean test} is a two-sided Welch test of the ten cores against the 400
controls, with a positive excess. The share asks whether many individual windows stand out; the mean
asks whether the cell as a whole moved, and can be swayed by one or two extreme windows. We call a
cell detectable when either passes and report which.

\paragraph{Learning rate.} One learning rate for every model turned out to be unfair: continued
training of Moirai at $10^{-4}$ makes its loss on unseen windows worse every epoch, and a model whose
training is breaking down fails to memorise for reasons unrelated to exposure. Each checkpoint
therefore uses the largest rate in $\{10^{-4}, 3\times10^{-5}, 10^{-5}\}$ that passes a rule fixed
before any planted run, applied to background windows only, and extended once after the TimesFM
runs (Appendix~\ref{app:lr}). TimesFM, whose
runs at the selected rate broke down part-way, uses $10^{-5}$, and Moirai large passed at no rate. The
resulting rates are in Table~\ref{tab:models}.

\subsection{The detectors respond to planted data}
\label{sec:positive}

On the planted 46M Chronos checkpoint, every loss-based detector separates exact copies from
never-planted windows at 320 exposures with AUROC of $0.99$ or more (Table~\ref{tab:detect}, Appendix~\ref{app:detect}). The
reference-calibrated score is the most sensitive, reaching \refExactTwenty\ at only 20 exposures, and
it is also the only one to pick up blends, at \refBlendTop\ after 320 exposures. The adaptation probe
responds the same way: windows planted 32 times need \probeTop$\times 10^{-3}$ of weight movement
against \probeHeld$\times 10^{-3}$ for never-planted windows, and from four copies upward the movement
falls as the dose rises; seed standard deviations are in Appendix~\ref{app:probe}. The scores and
the probe that were silent on the benchmark therefore do respond when memorised data is present.

\subsection{How many exposures memorisation needs}
\label{sec:family}

\begin{figure}[t]
\centering
\includegraphics[width=0.85\linewidth]{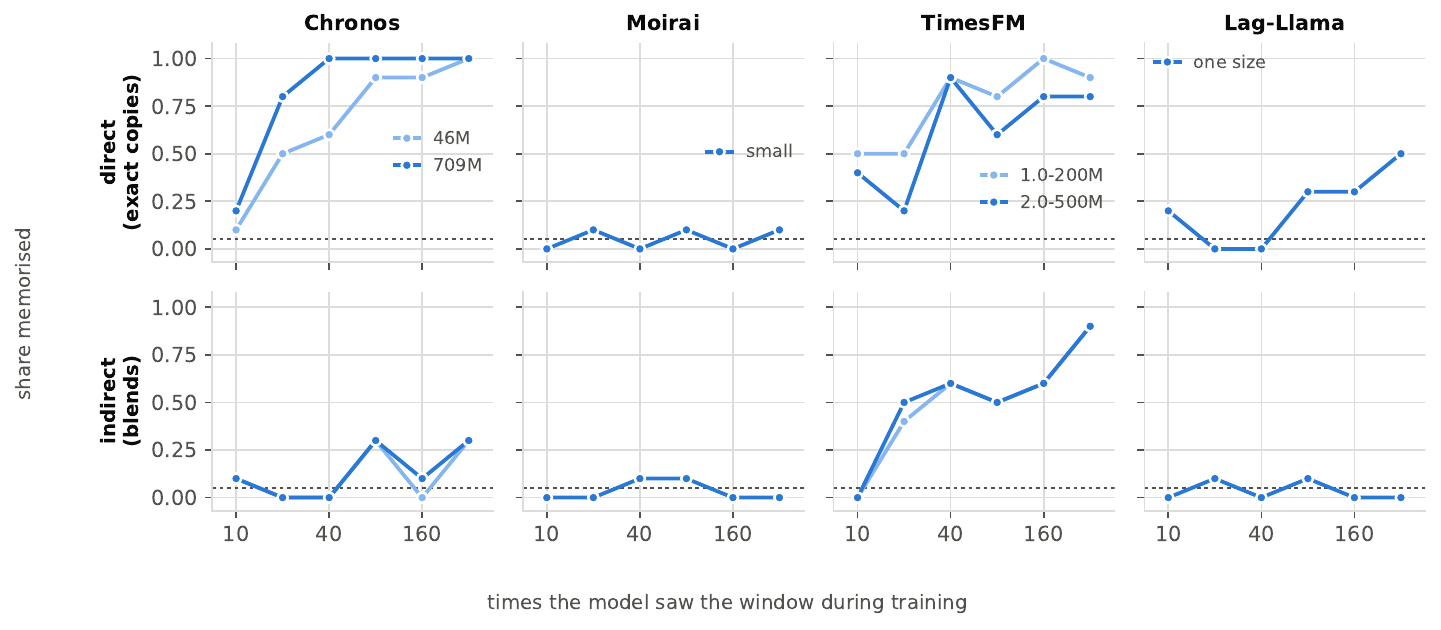}
\caption{Share of planted core windows that improve by more than the controls' 95th percentile
(dotted line: $0.05$, chance), against the number of times the model saw the window, for six
checkpoints from four families, each under its own continued-training schedule. Top row: exact copies
(direct); bottom row: blends (indirect). Each panel names its checkpoints. Moirai large is missing
because no learning rate in the grid trained it stably, and Lag-Llama exists in one size.}
\label{fig:family}
\end{figure}

\begin{table}[t]
\centering
\small
\caption{Smallest dose, in exposures, at which each form of planted leakage becomes detectable
(--- means at no dose up to 320). Each entry is the share test, with the mean test in brackets when
they differ. The last column comes from a second run in which every window was also scored after each
epoch, so doses from 1 to 10 exposures are observed (Bonferroni over all 180 cells).}
\label{tab:family}
\begin{tabular}{llrrrr}
\toprule
model & learning rate & exact & near-duplicate & blend & exact, scored each epoch \\
\midrule
Chronos 46M & $10^{-4}$ & 20 (40) & --- & --- &  \\
Chronos 709M & $10^{-4}$ & 20 & --- & --- & 10 (8) \\
TimesFM 1.0-200M & $10^{-5}$ & 10 (---) & 40 (---) & 20 (320) & 9 (---) \\
TimesFM 2.0-500M & $10^{-5}$ & 10 (---) & 40 (---) & 20 (320) &  \\
Moirai small & $10^{-5}$ & --- & --- & --- &  \\
Lag-Llama & $10^{-4}$ & 320 (80) & --- & --- (10)$^{\dagger}$ &  \\
\bottomrule
\end{tabular}

\\[2pt]
{\footnotesize $^{\dagger}$Under the stated rule this blend cell (10 exposures) is detectable: it
passes the mean test, with no core above the controls' 95th percentile, and nothing recurs at higher
doses.}
\end{table}

Figure~\ref{fig:family} and Table~\ref{tab:family} give the results for all six checkpoints. The
families behave very differently.

\emph{Exact copies} are memorised by every family except Moirai. Chronos 46M needs 20 to 40
exposures, the two tests marking the two ends, so its threshold under our rule is 20: at 20 exposures four of ten cores stand out and the
mean test gives $p = \pExactTwenty$, at 40 both pass (mean test $p = \pExactForty$; full grid in
Appendix~\ref{app:grid}). The 709M checkpoint is faster, with 80\% of its cores standing out at 20
exposures, in line with the general finding that larger models memorise more readily
\citep{carlini2023quantifying}. TimesFM is fastest under its own schedule: about half its cores stand out at the lowest dose
of the main run, 10 exposures. For TimesFM the share test passes at nearly every dose and the mean
test at none: at 320 exposures all ten cores improve on the controls, by $0.2$ to $4.3$, a spread that
leaves the mean of ten too uncertain to pass, so the share is the informative reading here. Lag-Llama
memorises slowly and incompletely: its mean improvement passes the corrected test at 80 exposures,
which is its threshold under our rule, but only at 320 do half its cores stand out. Moirai small shows
nothing at any dose. Because its learning rate is ten times smaller than Chronos's, we reran it at
$10^{-4}$, the rate that degrades its loss on unseen windows, and it still showed nothing. Its
likelihood might hide a trace, but rescoring a retrained copy of the planted run with TimesFM's
training loss, under the same tests, finds exact copies at no dose either (Appendix~\ref{app:moirai}).

\emph{Indirect leakage.} TimesFM is the only family with share-test detections for both
near-duplicates and blends: near-duplicates from 40 exposures and blends from 20 in the main run, and
from 16 and 24 when scored each epoch. Elsewhere the evidence is weaker but not absent. In Chronos,
neither form passes either test at any dose, but the reference-calibrated detector picks up blends at
320 exposures (AUROC \refBlendTop, Section~\ref{sec:positive}), so a moderate effect at that dose
cannot be ruled out. In Lag-Llama one blend cell passes the mean test at 10 exposures without
recurring. Moirai shows neither form under its own loss (for point losses, see Appendix~\ref{app:moirai}). TimesFM's point-forecast loss may be why: a forecaster that has seen a
near-copy can reproduce much of its horizon. We have not tested that explanation.

\emph{How often pretraining shows a window.} From the published Chronos training setup
\citep{ansari2024chronos} and the sampling assumptions stated in Appendix~\ref{app:exposure}, each day
of a USHCN record appears unmixed in about \expDay\ training crops and inside blends in about
\expBlend. A particular 576-day window, however, is presented exactly, as one aligned crop, only about
\expExact\ times, or once in \expOneIn. Our planted exposures are exact windows, so the two units
differ: by the planted unit, pretraining exposure is far below one; by days covered, it is about one,
in crops that overlap the window only partly. These figures are for Chronos. The published schedules
of the other families imply several passes over their corpora for TimesFM~1.0 and Moirai and about
one for Lag-Llama \citep{das2024timesfm,woo2024moirai,rasul2023lagllama}; Appendix~\ref{app:exposure}
gives the details.

\emph{How far below the threshold that is.} The main grid starts at 10 exposures, so for the two
checkpoints that memorised fastest, Chronos 709M and TimesFM 200M, we reran it scoring every window
after each epoch. A window planted once has then been seen 1, 2, \dots, 10 times. After a single
exposure, exact-copy cores are not detected: one of ten stands out for Chronos 709M and none for
TimesFM 200M, shares whose 95\% intervals reach \cpOneTen\ and \cpZeroTen. Exact copies first become
detectable at 8 to 10 exposures for Chronos 709M and 9 for TimesFM 200M. We therefore did not detect a
memorisation signal after one exposure in the two checkpoints tested at that dose. For Chronos 709M,
whose pretraining covers each day about once, the margin is about tenfold, not the twenty- to
fortyfold of Chronos 46M. TimesFM's schedule implies several passes over its corpus, so its
pretraining exposures sit closer to its threshold than Chronos's do; passes and planted exposures are
different units, so we do not put a number on that gap.

The per-epoch run also suggests that copies within one pass count for more than repeated passes
(Appendix~\ref{app:passes}).

\paragraph{Reading the thresholds.} Continued training on a small corpus makes memorisation easier
than pretraining does, in three ways. The planted copies are $19\%$ of the corpus, against something
like one window in ten million in a real pretraining corpus. The corpus is seen ten times, while Chronos
pretraining covers each day about once; for TimesFM, whose schedule makes several passes, this
difference is smaller. And the planted windows are the last thing the model learns, so nothing trained afterwards can
overwrite them, whereas a window seen once early in pretraining is followed by billions of other
tokens. All three suggest that our thresholds are lower than those that would apply in pretraining, but
reading them as floors is an untested extrapolation that supports at most the direction of the
correction, not its size. Model size pushes the other
way, as the two Chronos checkpoints show. In the units of the bound, the
mean excess of Chronos 46M's exact copies is \relExcessTen\% of cross-entropy at 10 exposures and
\relExcessTwenty\% at 20, so the bound of about $5\%$ sits below the dose at which exact copies produce a
double-digit gap; the two are different estimands, so this aligns scales rather than computes anything.

\section{Discussion, limitations, and conclusion}
\label{sec:discussion}

\paragraph{What the two halves say together.} On the one benchmark where membership can be verified,
the gap associated with a window being in the Chronos corpus is below about 5 to 6\% of token
cross-entropy, and models whose published corpora do not list those windows show the same gap. The
planted experiments are consistent with this: Chronos pretraining shows each day of these records
about once and a whole window almost never, and in the two checkpoints we tested at a single exposure
we detected no memorisation signal. They also show that the margin is not universal. How many
exposures memorisation needs varies by an order of magnitude between checkpoints, and for TimesFM,
whose planted near-duplicates and blends both pass the share test, indirect leakage is a live concern
even when the evaluation series themselves were never trained on. Our result is a
non-detection in this setting, not a general rule: in language models a single replica of a test set
in pretraining measurably lowers cross-entropy \citep{schaeffer2026quantifying}, and retraining Moirai
without its benchmark's training splits worsens its scores \citep{aksu2024gifteval}. An audit of one
checkpoint does not transfer to another.

\paragraph{Two habits for audits after the fact.} Both cost little. Check that a metric changes when
the model does before quoting it; MASE and WQL here would mostly have compared the data. And report a
model-free baseline beside any detector; ours showed that a comparison against post-release data would
have measured a decade of change in how rain falls.

\paragraph{Limitations.}
\begin{enumerate}
\item \textbf{One family and one variable carry the membership measurement.} A clean comparison needs
a corpus that contains a named subset of a larger archive, a same-period complement that can be
processed identically, and value-level verification. We found this for Chronos and daily
precipitation only (not for UCI electricity, whose corpus version cannot be reproduced from the raw
archive, nor for ERA5). The other families enter the membership comparison only as controls.
\item \textbf{The comparison is observational.} We bound a membership-associated gap, not an effect.
The difficulty adjustment and the cross-family control address the obvious confounds but cannot exclude
one we did not think of, and every adjusted figure is conditional on the adjustment's assumptions.
\item \textbf{Absence rests on listings.} Presence of \textbf{direct} is verified value by value, but
absence of \textbf{indirect} from the Chronos corpus, and absence of the gauges from the control
families' corpora, are established only at the level of dataset listings. Only \textbf{clean} is out of
the corpus by date. If a copy we missed exists, the contrast is weaker than we think and the numbers
would not show it.
\item \textbf{Analysis choices.} CE was made the primary metric after seeing MASE and WQL behave,
whereas the model-free check was fixed in advance; that check tests one descriptor at a time and cannot
resolve differences below an AUROC of about \mdeAuc; and a definition of difficulty we did not try
could give a different bound (Sections~\ref{sec:gate}, \ref{sec:metric} and~\ref{sec:headline}).
\item \textbf{The thresholds belong to particular training schedules.} They come from continued
training on a small, heavily contaminated corpus, with ten cores per cell, and learning rates differ
between families by a rule fixed in advance and extended once, for every run, after the TimesFM runs
failed (Appendix~\ref{app:lr}). The Chronos 46M grid was trained three times. The second run is the
one reported and the third, under the shared multi-family pipeline, reproduced its thresholds; the
first kept only summary means, which match the reported run's for exact copies to within 0.8 nats at
every dose (156.0 against 155.9 at 320 exposures), with the mean test first passing at the same dose
and neither indirect form passing in either run. Every other checkpoint's dose curve is a single run. Doses below ten exposures were observed for two checkpoints only, the two
tests disagree for TimesFM, where we rely on the share test, and the pretraining exposure estimate rests
on sampler assumptions stated in Appendix~\ref{app:exposure}.
\end{enumerate}

\paragraph{Reproducibility.} The supplementary file \texttt{benchmark\_windows.csv} lists every
evaluation window (station, group, first and last day, kept or not); Appendix~\ref{app:impl} gives the
implementation details. Every table and number in the text is generated by one script from the saved
per-window results. \codeavailability

\paragraph{Conclusion.} For the model whose corpus we could audit, the membership-associated gap is
below about 5 to 6\% of cross-entropy, and we detected no memorisation after one exposure in the two
checkpoints tested at that dose. Other checkpoints memorise faster or learn from related series, so the
audit has to be repeated per family, and general familiarity with a domain has to be handled when an
evaluation set is built.

\ifarxiv
\subsubsection*{Acknowledgments}
This work was carried out at Unilever's Horizon3 AI Labs in Toronto, Canada. We acknowledge the
financial support for this work from Mitacs. AI tools were used to help edit the text and write the
analysis code. The authors take full responsibility for the content.
\fi

\bibliographystyle{tmlr}
\bibliography{refs}

\begin{thebibliography}{21}
\providecommand{\natexlab}[1]{#1}
\providecommand{\url}[1]{\texttt{#1}}
\expandafter\ifx\csname urlstyle\endcsname\relax
  \providecommand{\doi}[1]{doi: #1}\else
  \providecommand{\doi}{doi: \begingroup \urlstyle{rm}\Url}\fi

\bibitem[Aksu et~al.(2024)Aksu, Woo, Liu, Liu, Liu, Savarese, Xiong, and
  Sahoo]{aksu2024gifteval}
Taha Aksu, Gerald Woo, Juncheng Liu, Xu~Liu, Chenghao Liu, Silvio Savarese,
  Caiming Xiong, and Doyen Sahoo.
\newblock {GIFT-Eval}: A benchmark for general time series forecasting model
  evaluation, 2024.
\newblock URL \url{https://arxiv.org/abs/2410.10393}.

\bibitem[Ansari et~al.(2024)Ansari, Stella, Turkmen, Zhang, Mercado, Shen,
  Shchur, Rangapuram, Pineda~Arango, Kapoor, Zschiegner, Maddix, Wang, Mahoney,
  Torkkola, Wilson, Bohlke-Schneider, and Wang]{ansari2024chronos}
Abdul~Fatir Ansari, Lorenzo Stella, Caner Turkmen, Xiyuan Zhang, Pedro Mercado,
  Huibin Shen, Oleksandr Shchur, Syama~Sundar Rangapuram, Sebastian
  Pineda~Arango, Shubham Kapoor, Jasper Zschiegner, Danielle~C. Maddix, Hao
  Wang, Michael~W. Mahoney, Kari Torkkola, Andrew~Gordon Wilson, Michael
  Bohlke-Schneider, and Yuyang Wang.
\newblock {C}hronos: Learning the language of time series.
\newblock \emph{Transactions on Machine Learning Research}, 2024.
\newblock URL \url{https://openreview.net/forum?id=gerNCVqqtR}.

\bibitem[Cameron et~al.(2008)Cameron, Gelbach, and
  Miller]{cameron2008bootstrap}
A.~Colin Cameron, Jonah~B. Gelbach, and Douglas~L. Miller.
\newblock Bootstrap-based improvements for inference with clustered errors.
\newblock \emph{Review of Economics and Statistics}, 90\penalty0 (3):\penalty0
  414--427, 2008.
\newblock \doi{10.1162/rest.90.3.414}.

\bibitem[Carlini et~al.(2023)Carlini, Ippolito, Jagielski, Lee, Tram{\`e}r, and
  Zhang]{carlini2023quantifying}
Nicholas Carlini, Daphne Ippolito, Matthew Jagielski, Katherine Lee, Florian
  Tram{\`e}r, and Chiyuan Zhang.
\newblock Quantifying memorization across neural language models.
\newblock In \emph{International Conference on Learning Representations
  (ICLR)}, 2023.
\newblock URL \url{https://arxiv.org/abs/2202.07646}.

\bibitem[Das et~al.(2024)Das, Kong, Sen, and Zhou]{das2024timesfm}
Abhimanyu Das, Weihao Kong, Rajat Sen, and Yichen Zhou.
\newblock A decoder-only foundation model for time-series forecasting.
\newblock In \emph{Proceedings of the 41st International Conference on Machine
  Learning}, volume 235 of \emph{Proceedings of Machine Learning Research},
  pp.\  10148--10167. PMLR, 2024.
\newblock URL \url{https://proceedings.mlr.press/v235/das24c.html}.

\bibitem[Das et~al.(2025)Das, Zhang, and Tram{\`e}r]{das2025blind}
Debeshee Das, Jie Zhang, and Florian Tram{\`e}r.
\newblock Blind baselines beat membership inference attacks for foundation
  models.
\newblock In \emph{2025 IEEE Security and Privacy Workshops (SPW)}, pp.\
  118--125. IEEE, 2025.
\newblock \doi{10.1109/SPW67851.2025.00016}.
\newblock URL \url{https://arxiv.org/abs/2406.16201}.

\bibitem[Duan et~al.(2024)Duan, Suri, Mireshghallah, Min, Shi, Zettlemoyer,
  Tsvetkov, Choi, Evans, and Hajishirzi]{duan2024membership}
Michael Duan, Anshuman Suri, Niloofar Mireshghallah, Sewon Min, Weijia Shi,
  Luke Zettlemoyer, Yulia Tsvetkov, Yejin Choi, David Evans, and Hannaneh
  Hajishirzi.
\newblock Do membership inference attacks work on large language models?
\newblock In \emph{Conference on Language Modeling (COLM)}, 2024.
\newblock URL \url{https://arxiv.org/abs/2402.07841}.

\bibitem[Gneiting \& Raftery(2007)Gneiting and Raftery]{gneiting2007strictly}
Tilmann Gneiting and Adrian~E. Raftery.
\newblock Strictly proper scoring rules, prediction, and estimation.
\newblock \emph{Journal of the American Statistical Association}, 102\penalty0
  (477):\penalty0 359--378, 2007.
\newblock \doi{10.1198/016214506000001437}.

\bibitem[Hanley \& McNeil(1982)Hanley and McNeil]{hanley1982meaning}
James~A. Hanley and Barbara~J. McNeil.
\newblock The meaning and use of the area under a receiver operating
  characteristic ({ROC}) curve.
\newblock \emph{Radiology}, 143\penalty0 (1):\penalty0 29--36, 1982.
\newblock \doi{10.1148/radiology.143.1.7063747}.

\bibitem[Hersbach et~al.(2020)Hersbach, Bell, Berrisford, Hirahara,
  Hor{\'a}nyi, Mu{\~n}oz-Sabater, Nicolas, Peubey, Radu, Schepers, Simmons,
  Soci, Abdalla, Abellan, Balsamo, Bechtold, Biavati, Bidlot, Bonavita,
  De~Chiara, Dahlgren, Dee, Diamantakis, Dragani, Flemming, Forbes, Fuentes,
  Geer, Haimberger, Healy, Hogan, H{\'o}lm, Janiskov{\'a}, Keeley, Laloyaux,
  Lopez, Lupu, Radnoti, de~Rosnay, Rozum, Vamborg, Villaume, and
  Th{\'e}paut]{hersbach2020era5}
Hans Hersbach, Bill Bell, Paul Berrisford, Shoji Hirahara, Andr{\'a}s
  Hor{\'a}nyi, Joaqu{\'i}n Mu{\~n}oz-Sabater, Julien Nicolas, Carole Peubey,
  Raluca Radu, Dinand Schepers, Adrian Simmons, Cornel Soci, Saleh Abdalla,
  Xavier Abellan, Gianpaolo Balsamo, Peter Bechtold, Gionata Biavati, Jean
  Bidlot, Massimo Bonavita, Giovanna De~Chiara, Per Dahlgren, Dick Dee, Michail
  Diamantakis, Rossana Dragani, Johannes Flemming, Richard Forbes, Manuel
  Fuentes, Alan Geer, Leo Haimberger, Sean Healy, Robin~J. Hogan, El{\'i}as
  H{\'o}lm, Marta Janiskov{\'a}, Sarah Keeley, Patrick Laloyaux, Philippe
  Lopez, Cristina Lupu, Gabor Radnoti, Patricia de~Rosnay, Iryna Rozum, Freja
  Vamborg, Sebastien Villaume, and Jean-No{\"e}l Th{\'e}paut.
\newblock The {ERA}5 global reanalysis.
\newblock \emph{Quarterly Journal of the Royal Meteorological Society},
  146\penalty0 (730):\penalty0 1999--2049, 2020.
\newblock \doi{10.1002/qj.3803}.

\bibitem[Hyndman \& Koehler(2006)Hyndman and Koehler]{hyndman2006another}
Rob~J. Hyndman and Anne~B. Koehler.
\newblock Another look at measures of forecast accuracy.
\newblock \emph{International Journal of Forecasting}, 22\penalty0
  (4):\penalty0 679--688, 2006.
\newblock \doi{10.1016/j.ijforecast.2006.03.001}.

\bibitem[Li et~al.(2026)Li, Xie, Shen, Li, Chen, Zhang, Fu, Sun, Ren, and
  Liu]{li2026tsfmaudit}
Hongkai Li, Shifeng Xie, Lefei Shen, Zhuo Li, Mouxiang Chen, Xiaobin Zhang, Han
  Fu, Jianling Sun, Xiaoxue Ren, and Chenghao Liu.
\newblock {TSFMA}udit: Data contamination auditing in forecasting time series
  foundation models, 2026.
\newblock URL \url{https://arxiv.org/abs/2605.26161}.

\bibitem[Magar \& Schwartz(2022)Magar and Schwartz]{magar2022contamination}
Inbal Magar and Roy Schwartz.
\newblock Data contamination: From memorization to exploitation.
\newblock In \emph{Proceedings of the 60th Annual Meeting of the Association
  for Computational Linguistics (Volume 2: Short Papers)}, pp.\  157--165,
  Dublin, Ireland, 2022. Association for Computational Linguistics.
\newblock \doi{10.18653/v1/2022.acl-short.18}.
\newblock URL \url{https://aclanthology.org/2022.acl-short.18/}.

\bibitem[Menne et~al.(2009)Menne, Williams, and Vose]{menne2009ushcn}
Matthew~J. Menne, Claude~N. Williams, Jr., and Russell~S. Vose.
\newblock The {U.S.} historical climatology network monthly temperature data,
  version 2.
\newblock \emph{Bulletin of the American Meteorological Society}, 90\penalty0
  (7):\penalty0 993--1008, 2009.
\newblock \doi{10.1175/2008BAMS2613.1}.

\bibitem[Menne et~al.(2012)Menne, Durre, Vose, Gleason, and
  Houston]{menne2012ghcn}
Matthew~J. Menne, Imke Durre, Russell~S. Vose, Byron~E. Gleason, and Tamara~G.
  Houston.
\newblock An overview of the global historical climatology network-daily
  database.
\newblock \emph{Journal of Atmospheric and Oceanic Technology}, 29\penalty0
  (7):\penalty0 897--910, 2012.
\newblock \doi{10.1175/JTECH-D-11-00103.1}.

\bibitem[Meyer et~al.(2025)Meyer, Kaltenpoth, Zalipski, and
  M{\"u}ller]{meyer2025rethinking}
Marcel Meyer, Sascha Kaltenpoth, Kevin Zalipski, and Oliver M{\"u}ller.
\newblock Rethinking evaluation in the era of time series foundation models:
  {(Un)known} information leakage challenges, 2025.
\newblock URL \url{https://arxiv.org/abs/2510.13654v3}.
\newblock Version 3, February 2026.

\bibitem[Rasul et~al.(2023)Rasul, Ashok, et~al.]{rasul2023lagllama}
Kashif Rasul, Arjun Ashok, et~al.
\newblock Lag-{L}lama: Towards foundation models for probabilistic time series
  forecasting, 2023.
\newblock URL \url{https://arxiv.org/abs/2310.08278}.

\bibitem[Schaeffer et~al.(2026)Schaeffer, Kazdan, Abbasi, Liu, Miranda, Ahmed,
  Berez, Puri, Biderman, Mireshghallah, and Koyejo]{schaeffer2026quantifying}
Rylan Schaeffer, Joshua Kazdan, Baber Abbasi, Ken~Ziyu Liu, Brando Miranda,
  Ahmed Ahmed, Fazl Berez, Abhay Puri, Stella Biderman, Niloofar Mireshghallah,
  and Sanmi Koyejo.
\newblock Quantifying the effect of test set contamination on generative
  evaluations, 2026.
\newblock URL \url{https://arxiv.org/abs/2601.04301}.

\bibitem[Shi et~al.(2024)Shi, Ajith, Xia, Huang, Liu, Blevins, Chen, and
  Zettlemoyer]{shi2024detecting}
Weijia Shi, Anirudh Ajith, Mengzhou Xia, Yangsibo Huang, Daogao Liu, Terra
  Blevins, Danqi Chen, and Luke Zettlemoyer.
\newblock Detecting pretraining data from large language models.
\newblock In \emph{International Conference on Learning Representations
  (ICLR)}, 2024.
\newblock URL \url{https://arxiv.org/abs/2310.16789}.

\bibitem[Shokri et~al.(2017)Shokri, Stronati, Song, and
  Shmatikov]{shokri2017membership}
Reza Shokri, Marco Stronati, Congzheng Song, and Vitaly Shmatikov.
\newblock Membership inference attacks against machine learning models.
\newblock In \emph{2017 IEEE Symposium on Security and Privacy (SP)}, pp.\
  3--18. IEEE, 2017.
\newblock \doi{10.1109/SP.2017.41}.

\bibitem[Woo et~al.(2024)Woo, Liu, Kumar, Xiong, Savarese, and
  Sahoo]{woo2024moirai}
Gerald Woo, Chenghao Liu, Akshat Kumar, Caiming Xiong, Silvio Savarese, and
  Doyen Sahoo.
\newblock Unified training of universal time series forecasting transformers.
\newblock In \emph{Proceedings of the 41st International Conference on Machine
  Learning}, volume 235 of \emph{Proceedings of Machine Learning Research},
  pp.\  53140--53164. PMLR, 2024.
\newblock URL \url{https://proceedings.mlr.press/v235/woo24a.html}.

\end{thebibliography}

\appendix
\section{Why MASE and WQL are nearly insensitive to the model here}
\label{app:metrics}

Table~\ref{tab:app-errors} gives mean error by group for every metric we computed, over the kept
windows. MASE divides the mean absolute error of the median of 100 sampled forecasts by the mean
absolute day-to-day change in the context \citep{hyndman2006another}. WQL averages the pinball loss
over the nine deciles and divides by the total observed rain; with the convention that carries a
factor of 2, an all-zero forecast scores exactly $1$. Scaled CRPS is the sample estimate
$\mathbb{E}|X-y| - \tfrac12\mathbb{E}|X-X'|$ \citep{gneiting2007strictly}, divided by the MASE
denominator.

\begin{table}[H]
\centering
\small
\caption{Mean error by group and checkpoint. CE is summed over 65 tokens, in nats; the others are
ratios. Lower is better. Gaps in the text are computed from unrounded per-window values, so they can
differ in the last digit from differences of these rounded means.}
\label{tab:app-errors}
\begin{tabular}{llrrrr}
\toprule
checkpoint & group & CE & MASE & WQL & sCRPS \\
\midrule
\multirow{3}{*}{20M} & direct & 141.44 & 0.6802 & 0.9984 & 0.6668 \\
 & indirect & 136.92 & 0.6466 & 0.9990 & 0.6314 \\
 & clean & 126.70 & 0.6263 & 0.9990 & 0.6112 \\
\midrule
\multirow{3}{*}{46M} & direct & 139.69 & 0.6802 & 0.9982 & 0.6644 \\
 & indirect & 134.65 & 0.6465 & 0.9977 & 0.6288 \\
 & clean & 124.43 & 0.6265 & 0.9980 & 0.6106 \\
\midrule
\multirow{3}{*}{201M} & direct & 133.34 & 0.6802 & 0.9930 & 0.6577 \\
 & indirect & 128.68 & 0.6466 & 0.9938 & 0.6248 \\
 & clean & 119.07 & 0.6264 & 0.9922 & 0.6006 \\
\midrule
\multirow{3}{*}{709M} & direct & 130.73 & 0.6802 & 0.9898 & 0.6496 \\
 & indirect & 126.34 & 0.6465 & 0.9900 & 0.6179 \\
 & clean & 116.65 & 0.6263 & 0.9873 & 0.5967 \\
\bottomrule
\end{tabular}

\end{table}

The MASE column barely moves across a 35-fold change in model size; per window, MASE is exactly
identical between the 20M model and each larger one on \maseSameLo--\maseSameHi\% of windows. With
about \dryAll\% of days dry, the median forecast is zero on most horizon days and the error reduces to
$\mathrm{mean}_t|y_t|$, which does not involve the model: an all-zero forecast implies a pooled mean
MASE of \maseZero, and the 46M model's measured value is \maseSmall. For WQL, an all-zero forecast
gives a pinball loss of $2q\sum_t y_t$ at level $q$, and averaging over the deciles gives exactly
$\sum_t y_t$, the normaliser; the measured median over windows is \wqlMedian\ and \wqlNear\% of windows
lie within $[0.99, 1.01]$. Scaled CRPS responds to the model but correlates with the difficulty
measure $d$ at \corrDScrps, so adjusting for $d$ leaves almost nothing of it. Adjusted in the same way
as CE (Section~\ref{sec:confound}), its one-sided 97.5\% bound on the membership-associated gap is
\scrpsAdjBySize\% of the \textbf{indirect} mean for the 20M, 46M, 201M and 709M models, against
\scrpsRawLo--\scrpsRawHi\% before adjustment. CE correlates with $d$
at \corrDCe.

\section{Resampling nearby gauges together}
\label{app:blockboot}

Resampling stations treats different gauges as independent, but nearby gauges record the same storms:
the median \textbf{indirect} station is \distMedian\,km from a corpus station. As a sensitivity check
we repeat every interval with whole geographic cells as the resampling unit, so that gauges sharing
weather are drawn together: \nTwoDegUnits\ cells of $2^\circ$ by $2^\circ$, and \nFiveDegUnits\ of
$5^\circ$ by $5^\circ$, against \nStationUnits\ stations. Table~\ref{tab:blockboot} gives the bounds.
The $2^\circ$ cells widen the intervals and raise the adjusted bound to \blockTwoLo--\blockTwoHi\%;
the $5^\circ$ cells give \blockFiveLo--\blockFiveHi\%, below the station bounds in most rows. That
is the opposite of what coarser units should do, and we read it as noise: with only \nFiveDegUnits\
units a cluster bootstrap is rough and tends to understate the spread \citep{cameron2008bootstrap}.
We rely on the $2^\circ$ result, and report the station-level figure as the main result with the
$2^\circ$ figure beside it as the more conservative one.

\begin{table}[H]
\centering
\small
\caption{The bound on the membership-associated gap in CE under three resampling units. Station
columns are the bounds of Table~\ref{tab:bound}; cell columns resample whole latitude--longitude cells,
so that nearby gauges are drawn together. One-sided 97.5\%, percentage of the \textbf{indirect} mean.}
\label{tab:blockboot}
\begin{tabular}{lrrrcrrr}
\toprule
& \multicolumn{3}{c}{raw} & & \multicolumn{3}{c}{adjusted for difficulty} \\
\cmidrule(lr){2-4}\cmidrule(lr){6-8}
checkpoint & stations & $2^\circ$ cells & $5^\circ$ cells & & stations & $2^\circ$ cells & $5^\circ$ cells \\
\midrule
20M & 4.6\% & 4.9\% & 4.2\% & & 5.2\% & 5.7\% & 5.1\% \\
46M & 4.2\% & 4.4\% & 3.8\% & & 4.8\% & 5.2\% & 4.8\% \\
201M & 4.4\% & 4.5\% & 3.6\% & & 4.9\% & 5.3\% & 4.9\% \\
709M & 4.4\% & 4.5\% & 3.8\% & & 5.0\% & 5.7\% & 4.8\% \\
\bottomrule
\end{tabular}

\end{table}

\section{Robustness to the definition of difficulty}
\label{app:robust}

\begin{table}[H]
\centering
\small
\caption{The adjusted bound on CE, \textbf{direct} versus \textbf{indirect}, under four model-free
definitions of difficulty. Correlation and $R^2$ are against CE at 46M and are stable across sizes.
Every bound is conditional on the adjustment's assumptions (linearity in the covariate, overlap of the
groups, no further confounder). The dry-day share explains nearly all of CE by itself, so those
assumptions carry the most weight for it. The reported definition gives the loosest bound of the four.}
\label{tab:app-robust}
\begin{tabular}{lrrrrrr}
\toprule
covariate & corr.\ with CE & $R^2$ & 20M & 46M & 201M & 709M \\
\midrule
$d$ (reported) & $+$0.441 & 0.19 & \textbf{5.2\%} & \textbf{4.8\%} & \textbf{4.9\%} & \textbf{5.0\%} \\
horizon s.d. & $+$0.562 & 0.32 & 2.5\% & 2.1\% & 2.1\% & 2.2\% \\
context variability & $+$0.509 & 0.26 & 1.4\% & 0.8\% & 0.9\% & 1.0\% \\
dry-day share & $-$0.972 & 0.94 & 3.3\% & 3.0\% & 3.1\% & 3.2\% \\
\bottomrule
\end{tabular}

\end{table}

\section{The cross-family comparison}
\label{app:did}

\begin{figure}[H]
\centering
\includegraphics[width=0.8\linewidth]{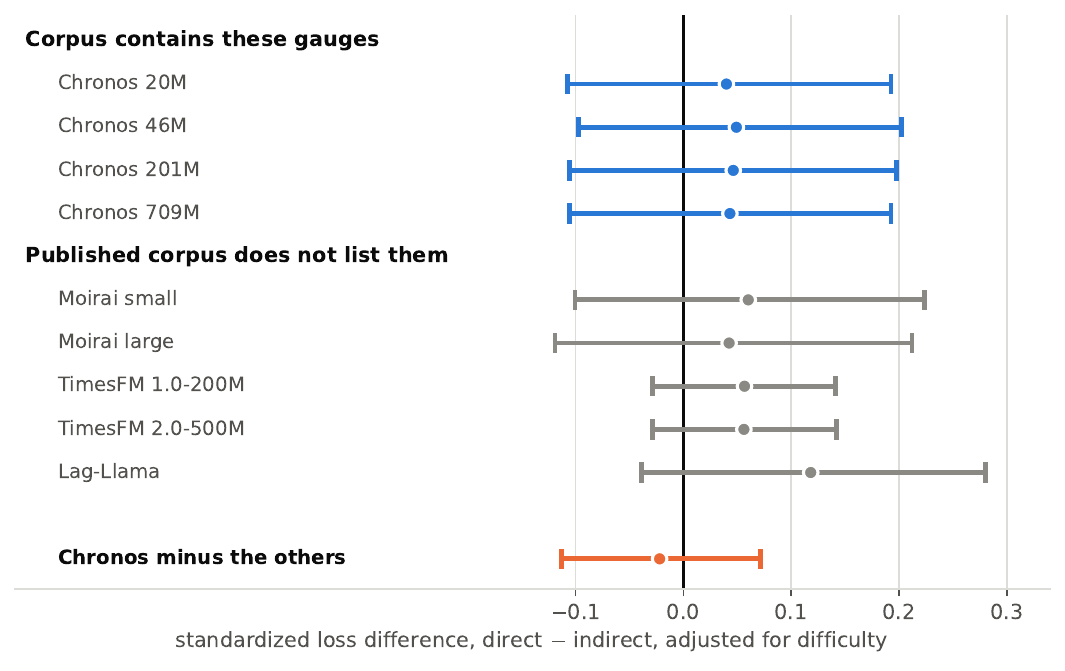}
\caption{\textbf{direct} minus \textbf{indirect} loss for all nine checkpoints, adjusted for
difficulty, in standard deviations of each model's per-window loss, with 95\% station-bootstrap
intervals. Blue: Chronos, whose corpus contains the \textbf{direct} gauges. Grey: checkpoints whose
published corpora do not list them. Orange: the Chronos average minus the average of the others,
adjusted, on the same bootstrap resamples (the raw version is in Table~\ref{tab:did}).}
\label{fig:step1}
\end{figure}

Table~\ref{tab:did} and Figure~\ref{fig:step1} give the per-checkpoint gaps behind
Section~\ref{sec:did}, before and after adjusting for difficulty.

\begin{table}[H]
\centering
\small
\caption{\textbf{direct} minus \textbf{indirect} loss, in standard deviations of each model's
per-window loss, before and after adjusting for difficulty; 95\% station-bootstrap intervals. Only
Chronos has the \textbf{direct} gauges in its corpus; for the others, absence is established at the
level of dataset listings. The last row is the average of the four Chronos checkpoints minus the average
of the other five, for each column, computed on the same bootstrap resamples.}
\label{tab:did}
\begin{tabular}{llrr}
\toprule
model & USHCN in corpus & raw gap & adjusted gap \\
\midrule
Chronos 20M & yes & $+0.070\ [-0.101,\ 0.239]$ & $+0.040\ [-0.108,\ 0.192]$ \\
Chronos 46M & yes & $+0.079\ [-0.084,\ 0.243]$ & $+0.049\ [-0.097,\ 0.202]$ \\
Chronos 201M & yes & $+0.076\ [-0.092,\ 0.245]$ & $+0.046\ [-0.106,\ 0.198]$ \\
Chronos 709M & yes & $+0.074\ [-0.096,\ 0.243]$ & $+0.043\ [-0.106,\ 0.193]$ \\
Moirai small & no & $+0.076\ [-0.093,\ 0.246]$ & $+0.060\ [-0.101,\ 0.224]$ \\
Moirai large & no & $+0.060\ [-0.107,\ 0.232]$ & $+0.042\ [-0.119,\ 0.212]$ \\
TimesFM 200M & no & $+0.113\ [-0.044,\ 0.271]$ & $+0.057\ [-0.029,\ 0.141]$ \\
TimesFM 500M & no & $+0.113\ [-0.042,\ 0.273]$ & $+0.056\ [-0.029,\ 0.142]$ \\
Lag-Llama & no & $+0.141\ [-0.028,\ 0.309]$ & $+0.118\ [-0.039,\ 0.280]$ \\
\midrule
Chronos minus the others & & $-0.026\ [-0.122,\ 0.070]$ & $-0.022\ [-0.113,\ 0.072]$ \\
\bottomrule
\end{tabular}

\end{table}

\section{Membership detectors}
\label{app:detect}

Table~\ref{tab:detect} gives the AUROCs behind Sections~\ref{sec:detectors} and~\ref{sec:positive}.

\begin{table}[H]
\centering
\small
\caption{Loss-based membership detectors, AUROC. Left: \textbf{direct} against \textbf{indirect} on
the benchmark, with 95\% station-bootstrap intervals. Right: planted cores against never-planted
controls on the planted 46M checkpoint (Section~\ref{sec:inject-design}), for exact copies at 20 and
320 exposures and blends at 320. The reference-calibrated score uses Moirai small as reference on the
benchmark and the released 46M checkpoint on planted data.}
\label{tab:detect}
\begin{tabular}{lrrrrr}
\toprule
& \multicolumn{2}{c}{benchmark: direct vs.\ indirect} & \multicolumn{3}{c}{planted vs.\ never planted (46M)} \\
\cmidrule(lr){2-3}\cmidrule(lr){4-6}
score & 46M & 709M & exact, 20 & exact, 320 & blend, 320 \\
\midrule
loss & 0.481 $[0.43,\ 0.53]$ & 0.483 $[0.44,\ 0.53]$ & 0.52 & 1.00 & 0.37 \\
Min-20\% & 0.491 $[0.44,\ 0.54]$ & 0.499 $[0.45,\ 0.55]$ & 0.59 & 0.99 & 0.47 \\
difficulty-calibrated & 0.486 $[0.44,\ 0.53]$ & 0.488 $[0.44,\ 0.54]$ & 0.47 & 0.99 & 0.45 \\
reference-calibrated & 0.488 $[0.44,\ 0.53]$ & 0.491 $[0.44,\ 0.54]$ & 0.89 & 1.00 & 0.78 \\
\bottomrule
\end{tabular}

\end{table}

\section{Choosing the learning rate}
\label{app:lr}

One learning rate for every model turned out to be unfair: continued
training of Moirai at $10^{-4}$ makes its loss on unseen windows worse every epoch, and a model whose
training is breaking down fails to memorise for reasons unrelated to exposure. Before any planted run
we fixed a rule: each checkpoint uses the largest rate in $\{10^{-4}, 3\times10^{-5}, 10^{-5}\}$ at
which two epochs on 3{,}000 background windows leave the loss on 200 other background windows no more
than $1\%$ above where it started. The rule sees only background windows. The two-epoch check proved
too short for TimesFM: at the $3\times10^{-5}$ it selected, the ten-epoch runs broke down part-way,
the average training loss of single epochs jumping from about 1 to several hundred. We added a second
condition, applied to every run: no epoch's average training loss may exceed three times the first
epoch's. This condition was added after seeing the TimesFM runs, though it is applied to every run.
Only the TimesFM runs failed it; we discarded them and used $10^{-5}$. Moirai large passed at
no rate and is missing. The resulting rates are in Table~\ref{tab:models}. All runs use a 100-step
warm-up and gradient clipping.

\section{The full planted-data grid for Chronos 46M}
\label{app:grid}

\begin{table}[H]
\centering
\small
\caption{Chronos 46M, ten epochs, $k$ copies ($10k$ exposures). For each form and $k$: median and mean
improvement minus the control mean, in nats; share of cores above the control's 95th percentile; and
the two-sided Welch $p$-value against the 400 controls (Bonferroni threshold \bonf). With ten cores
the means are noisy: at $k{=}2$ two exact-copy cores gain \outlierA\ and \outlierB\ nats over the
controls, and the cell's standard error is \seCellTwenty\ nats. This is the run whose weights the probe
and the detectors of Section~\ref{sec:detect} use.}
\label{tab:app-grid}
\begin{tabular}{llrrrrrr}
\toprule
form & & $k{=}1$ & $k{=}2$ & $k{=}4$ & $k{=}8$ & $k{=}16$ & $k{=}32$ \\
\midrule
\multirow{4}{*}{exact} & median & $-1.0$ & $14.4$ & $17.1$ & $23.6$ & $72.4$ & $150.3$ \\
 & mean & $2.8$ & $21.8$ & $17.7$ & $32.1$ & $78.5$ & $155.9$ \\
 & share & 0.10 & 0.40 & 0.60 & 0.80 & 0.90 & 1.00 \\
 & $p$ & 0.39 & 0.036 & $4.9{\times}10^{-4}$ & 0.002 & $4.3{\times}10^{-4}$ & $1.8{\times}10^{-6}$ \\
\midrule
\multirow{4}{*}{near-duplicate} & median & $-4.7$ & $-2.6$ & $0.3$ & $1.3$ & $0.3$ & $2.4$ \\
 & mean & $0.1$ & $-1.8$ & $1.7$ & $3.5$ & $1.4$ & $7.6$ \\
 & share & 0.10 & 0.00 & 0.00 & 0.10 & 0.00 & 0.20 \\
 & $p$ & 0.98 & 0.31 & 0.50 & 0.36 & 0.62 & 0.24 \\
\midrule
\multirow{4}{*}{blend} & median & $-5.9$ & $-1.5$ & $-4.4$ & $-2.8$ & $-3.4$ & $3.6$ \\
 & mean & $-3.1$ & $-3.7$ & $-2.6$ & $4.5$ & $-2.2$ & $18.1$ \\
 & share & 0.10 & 0.00 & 0.00 & 0.20 & 0.00 & 0.30 \\
 & $p$ & 0.36 & 0.049 & 0.19 & 0.35 & 0.15 & 0.072 \\
\bottomrule
\end{tabular}

\end{table}

\begin{figure}[H]
\centering
\includegraphics[width=0.72\linewidth]{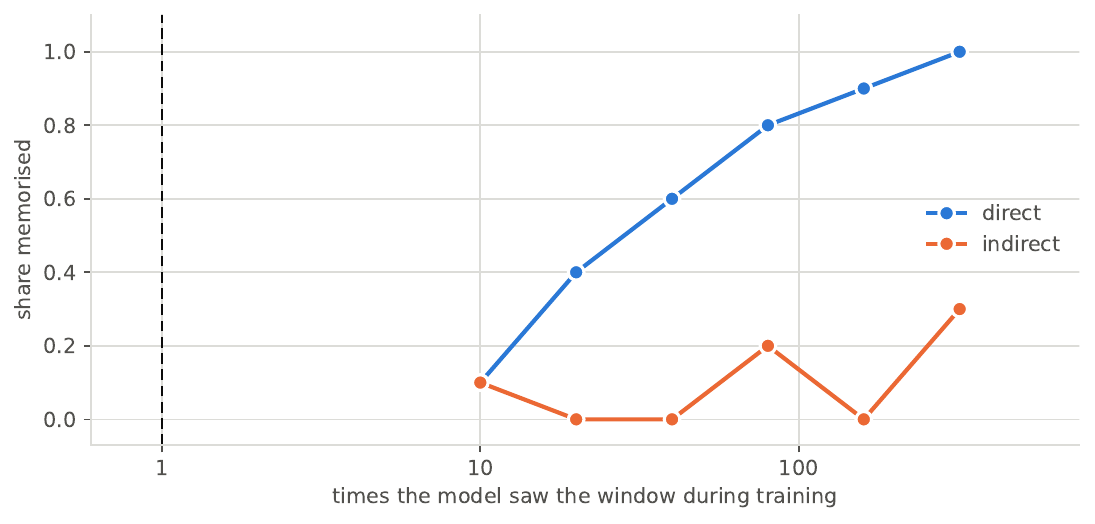}
\caption{Chronos 46M: share of planted cores standing out from the controls, exact copies against
blends. The dashed line marks one exposure. By the estimate of Appendix~\ref{app:exposure}, Chronos
pretraining covers each day of a USHCN record in about \expDay\ unmixed crops but presents a whole
window exactly far less than once, so the line is a reference point, not the pretraining dose in the
planted unit.}
\label{fig:app-inject}
\end{figure}

\section{Copies within a pass and repeated passes}
\label{app:passes}

The per-epoch run also bears on an assumption behind reporting doses in exposures: that exposures
accumulate the same way whether they come from more copies or more epochs. In Chronos 709M, eight
copies seen within the first epoch (8 exposures) are detectable, while one copy seen over ten epochs
(10 exposures) is not. The cores differ between those cells and so does the training state when they
are scored, so this suggests, rather than shows, that copies within one pass count for more than
repeated passes, which would matter for pretraining, where duplicated data appears several times
within its single pass.

\section{Moirai rescored with point losses}
\label{app:moirai}

Section~\ref{sec:family} leaves two readings of Moirai's null: it may memorise less, or its
mixture likelihood may not expose the trace to our tests. To check the second reading, we retrained the planted run, which reproduced its losses exactly, and rescored every window
with TimesFM's training loss, under the same share and mean tests as the main grid. Exact copies are
not detectable at any dose. We also scored the absolute error of the median forecast and the pinball
loss over the deciles, but on this data these barely respond to the model, for the same reason as MASE
and WQL (Section~\ref{sec:metric}), so their null says little. The one cell that passes under them, blends at
\moiraiPtHitDose\ exposures under the absolute error, has exactly \moiraiPtHitCores\ of ten cores
standing out and no support from the mean test ($p = \moiraiPtHitP$); 320 is the highest dose we
tested, so whether it would persist cannot be seen. Under TimesFM's loss no form passes at any dose. This
narrows the question rather than settles it, but it gives no sign that Moirai memorised exact copies
its likelihood failed to show.

\section{The adaptation probe}
\label{app:probe}

We fine-tune for 120 steps at batch size 8 and learning rate $5\times10^{-5}$, from the same starting
weights, on ten windows per group, over five seeds, and record the relative weight change over the
transformer matrices. Each seed draws a different ten windows from the group and a different
optimisation order, so the standard deviation reflects both. The unit scored is a group of ten windows. Table~\ref{tab:app-probe} gives the dose sweep on the planted 46M checkpoint and
the benchmark groups on the released one; both come from the same probe run.

\begin{table}[H]
\centering
\small
\caption{Relative weight movement $\lVert\Delta\theta\rVert/\lVert\theta\rVert$ ($\times 10^{-3}$),
mean $\pm$ standard deviation over five seeds; the unit is a group of ten windows. Less movement means
the model already fits the data.}
\label{tab:app-probe}
\begin{tabular}{lrlr}
\toprule
\multicolumn{2}{c}{planted checkpoint} & \multicolumn{2}{c}{released checkpoint} \\
\cmidrule(lr){1-2}\cmidrule(lr){3-4}
dose & movement & group & movement \\
\midrule
exact, $k{=}32$ & 2.680 $\pm$ 0.016 & direct & 7.384 $\pm$ 0.252 \\
exact, $k{=}16$ & 4.107 $\pm$ 0.024 & indirect & 7.115 $\pm$ 0.060 \\
exact, $k{=}8$ & 5.509 $\pm$ 0.051 & clean & 7.044 $\pm$ 0.189 \\
exact, $k{=}4$ & 6.349 $\pm$ 0.077 &  &  \\
exact, $k{=}2$ & 6.659 $\pm$ 0.019 &  &  \\
exact, $k{=}1$ & 6.505 $\pm$ 0.054 &  &  \\
never planted & 7.052 $\pm$ 0.115 &  &  \\
\bottomrule
\end{tabular}

\end{table}

At $k = 1$ the movement is \probeKone\ $\pm$ \probeSdKone, against \probeHeld\ $\pm$
\probeSdHeld\ for never-planted windows, so the two differ by several seed standard deviations. A
$k = 1$ window was seen ten times in a corpus of ten thousand, far more favourable to detection than
pretraining, so this sensitivity does not transfer to pretraining.

\section{How often pretraining shows a window}
\label{app:exposure}

The estimate uses the training setup published by \citet{ansari2024chronos} (their Section~5.2,
Table~3 and Algorithm~1): 200{,}000 steps of 256 sequences; each sequence a random crop of 576 steps
(512 context and 64 horizon) from a series drawn 9:1 from 10 million TSMixup augmentations and 1 million
KernelSynth series; each augmentation mixes $k \sim U\{1,2,3\}$ mean-scaled series with
Dirichlet($1.5$) weights over a length $l \sim U\{128, \dots, 2048\}$, each component taken from one of
the 28 training datasets chosen uniformly; USHCN contributes 6{,}090 series of mean length 38{,}653
days. Four assumptions are needed that the paper does not state: (A1) within a dataset the series and
the segment start are drawn uniformly; (A2) training draws are uniform over the augmentations; (A3) a
crop of an augmentation shorter than 576 steps contains all of it; (A4) the five USHCN variables are
treated alike.

Under these assumptions each augmentation is drawn $200{,}000 \times 256 \times 0.9 / 10^7 \approx
\drawsPerAug$ times. A fraction $\tfrac13 \cdot \tfrac1{28}$ of augmentations is a single unmixed USHCN
segment, about 119{,}000 of them. Averaging the share of an augmentation covered by one crop over $l$,
each day of a USHCN record falls inside about \expDay\ unmixed training crops, and inside about
\expBlend\ crops of augmentations in which it is blended with other series. A particular 576-day window
is presented exactly only if a crop starts on its first day within an unmixed augmentation long enough
to hold it; the expected number of such presentations is about \expExact, once in \expOneIn. The planted
experiments of Section~\ref{sec:detect} present exact windows, so their exposures correspond to the
second figure; the first figure counts crops that overlap a window only partly, a different unit.
The script \texttt{exposure\_estimate.py} reproduces these numbers.

\paragraph{Other families.} These figures are for Chronos, and the other families sample
differently. TimesFM~1.0 trains for 1.5 million steps of 4{,}096 windows \citep{das2024timesfm}, which
amounts to several passes over its corpus, and Moirai for 100{,}000 (small) to 1 million steps of 256
packed sequences \citep{woo2024moirai}, more than one pass on average but unevenly, because each
sub-dataset's sampling share is capped. The released Lag-Llama checkpoint stopped after 234 epochs of
128 batches, roughly one pass over its 352 million time steps, fewer or more depending on whether its
batch size was the 256 the paper reports or the 512 in its pretraining script \citep{rasul2023lagllama}.

\section{Implementation details}
\label{app:impl}

Value matching against the released \texttt{ushcn\_daily} rows converts the corpus's hundredths of an inch to
tenths of a millimetre and requires exact equality on all 576 days. Near-duplicates are
$0.99\,z + \sqrt{1 - 0.99^2}\,s$ for the standardised core $z$ and a phase-randomised surrogate $s$ with
the core's amplitude spectrum and rank-matched marginal, rescaled to the core's mean and standard
deviation; blends are $\lambda c + (1-\lambda) o$ with $\lambda \sim U(0.3, 0.7)$ and $o$ a random
background window. Continued training uses AdamW with PyTorch's default weight decay of $0.01$, batch
16, ten epochs, a 100-step linear warm-up, gradient-norm clipping at 1 and float32 weights, at the
rates in Table~\ref{tab:models}; the probe uses 120 steps, batch 8 and learning rate
$5\times10^{-5}$. Bootstrap intervals use 10{,}000 resamples with fixed seeds.

\end{document}